\def\year{2020}\relax
\documentclass[letterpaper]{article} 
\usepackage{aaai20}  
\usepackage{times}  
\usepackage{helvet} 
\usepackage{courier}  
\usepackage[hyphens]{url}  
\usepackage{graphicx} 
\usepackage{graphicx}  
\usepackage[utf8]{inputenc} 
\usepackage[T1]{fontenc}    
\usepackage{url}            
\usepackage{booktabs}       
\usepackage{amsfonts}       
\usepackage{nicefrac}       
\usepackage{microtype}      
\usepackage{subcaption}
\usepackage{boldline}
\usepackage{cuted}
\usepackage{capt-of}
\usepackage{tikz}
\usepackage{amsmath}
\usepackage{amssymb}
\usepackage{acronym}
\usepackage{verbatim}
\usepackage{xspace}
\usepackage{enumitem}
\usepackage{multirow}

\DeclareMathOperator*{\argmax}{arg\,max}

\newcommand{\citet}[1]{\citeauthor{#1} \citeyear{#1}}

\makeatletter
\DeclareRobustCommand\onedot{\futurelet\@let@token\@onedot}
\def\@onedot{\ifx\@let@token.\else.\null\fi\xspace}
\def\eg{\emph{e.g}\onedot} 
\def\ie{\emph{i.e}\onedot}

\makeatother

\acrodef{rl}[RL]{reinforcement learning}
\acrodef{eig}[EIG]{expected information gain}
\acrodef{cc}[CC]{Common Cause}
\acrodef{ce}[CE]{Common Effect}
\acrodef{cc3}[CC3]{Common Cause 3}
\acrodef{ce3}[CE3]{Common Effect 3}
\acrodef{cc4}[CC4]{Common Cause 4}
\acrodef{ce4}[CE4]{Common Effect 4}
\acrodef{mcmc}[MCMC]{Markov Chain Monte Carlo}
\acrodef{dqn}[DQN]{Deep Q-Network}
\acrodef{dqnpe}[DQN (PE)]{DQN with prioritized experience replay}
\acrodef{a2c}[A2C]{Advantage Actor-Critic}
\acrodef{trpo}[TRPO]{Trust Region Policy Optimization}
\acrodef{ppo}[PPO]{Proximal Policy Optimization}
\acrodef{maml}[MAML]{Model-Agnostic Meta-Learning}

\usetikzlibrary{calc,trees,positioning,arrows,chains,shapes.geometric,%
    decorations.pathreplacing,decorations.pathmorphing,shapes,%
    matrix,shapes.symbols}

\tikzset{
>=stealth',
  punktchain/.style={
    text width=3.5cm,
    minimum height=1em,
    text centered,
    on chain},
  line/.style={draw, thick, <-},
  element/.style={
    minimum width=1em,
    text width=8em,
    minimum height=2em,
    text centered,
    on chain},
  every join/.style={white, thick,shorten <=0.5pt},
  decoration={brace},
  tuborg/.style={decorate},
  tubnode/.style={midway, right=2pt},
}
\tikzset{mjefont/.style = {font=\normalsize}
    }

\begin{document}
\title{Theory-based Causal Transfer:\\Integrating Instance-level Induction and Abstract-level Structure Learning}

\author{
    Mark Edmonds,\textsuperscript{\rm 1,2}
    Xiaojian Ma,\textsuperscript{\rm 1}
    Siyuan Qi,\textsuperscript{\rm 1,2}
    Yixin Zhu,\textsuperscript{\rm 1,2}
    Hongjing Lu,\textsuperscript{\rm 3}
    Song-Chun Zhu\textsuperscript{\rm 1,2} \\
    \textsuperscript{\rm 1}UCLA Center for Vision, Cognition, Learning, and Autonomy \\
    \textsuperscript{\rm 2}International Center for AI and Robot Autonomy (CARA) \\
    \textsuperscript{\rm 3}UCLA Computational Vision and Learning (CVL) Lab \\
    \{markedmonds,maxiaojian,syqi,yixin.zhu,hongjing\}@ucla.edu, sczhu@stat.ucla.edu
}

\maketitle

\begin{abstract}
Learning transferable knowledge across similar but different settings is a fundamental component of generalized intelligence.
In this paper, we approach the transfer learning challenge from a causal theory perspective. 
Our agent is endowed with two basic yet general theories for transfer learning: (i) a task shares a common abstract structure that is invariant across domains, and (ii) the behavior of specific features of the environment remain constant across domains.
We adopt a Bayesian perspective of causal theory induction and use these theories to transfer knowledge between environments. Given these general theories, the goal is to train an agent by interactively exploring the problem space to (i) discover, form, and transfer useful \emph{abstract and structural} knowledge, and (ii) induce useful knowledge from the \emph{instance-level attributes} observed in the environment.
A hierarchy of Bayesian structures is used to model abstract-level structural causal knowledge, and an instance-level associative learning scheme learns which specific objects can be used to induce state changes through interaction. This model-learning scheme is then integrated with a model-based planner to achieve a task in the OpenLock environment, a virtual ``escape room'' with a complex hierarchy that requires agents to reason about an abstract, generalized causal structure.
We compare performances against a set of predominate model-free \ac{rl} algorithms. 
\ac{rl} agents showed poor ability transferring learned knowledge across different trials. Whereas the proposed model revealed similar performance trends as human learners, and more importantly, demonstrated transfer behavior across trials and learning situations.\footnote{The proposed algorithm and all baseline algorithms can be found on the first author's website.}
\end{abstract}

\section{Introduction}\label{sec:intro}

The ability of agents to learn and \emph{reuse} knowledge is a fundamental characteristic of general intelligence and is essential for agents to succeed in novel circumstances~\cite{legg2007universal}. Humans demonstrate a remarkable ability to transfer causal knowledge between environments governed by the same underlying mechanics, in spite of observational changes to the features of the environment~\cite{edmonds2018human}. Early psychological research framed causal understanding as learning stimulus-response relationships through observation in classical conditioning experimental paradigms~\cite{shanks1988associative,rescorla1972theory}. However, more recent studies show human understanding of causal mechanisms in the distal world is more complex than covariation between observed (perceptual) variables~\cite{holyoak2011causal}; \eg, humans \emph{explore} and \emph{experiment} with dynamic physical scenarios to refine causal hypotheses~\cite{bramley2018intuitive,stahl2015observing}.

\begin{figure}[t!]
    \centering
     \includegraphics[width=\linewidth]{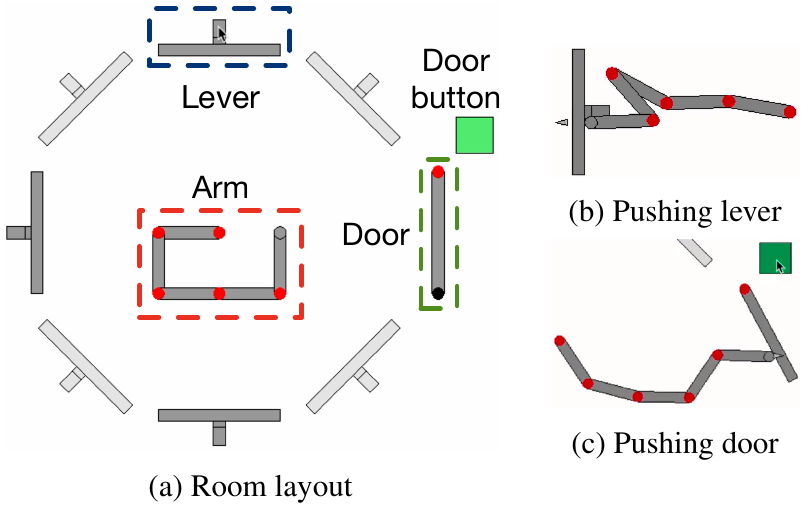}
    \caption{(a) Starting configuration of a 3-lever OpenLock room. The arm can interact with levers by either \emph{pushing} outward or \emph{pulling} inward, achieved by clicking either the outer or inner regions of the levers' radial tracks, respectively. Light gray levers are always locked; however, this is unknown to agents. The door can be pushed only after being unlocked. The green button serves as the mechanism to push on the door. The black circle on the door indicates whether or not the door is unlocked; locked if present, unlocked if absent. (b) Pushing on a lever. (c) Opening the door.}
    \label{fig:starting-config-and-pushing}
\end{figure}

Since the associative account, researchers have demonstrated that humans uncover causal relationships through the discovery of abstract causal structure~\cite{waldmann1992predictive} and causal strength~\cite{cheng1997covariation}. Simultaneously, causal graphical models and Bayesian statistical inference have been developed to provide a general representational framework for how causal structure and strength are discovered~\cite{griffiths2005structure,griffiths2009theory,tenenbaum2006theory,bramley2015conservative,bramley2017formalizing,holyoak2011causal}. Under such a framework, causal connections encode a structural model of the world. States represent some status in the world, and connections between states imply the presence of a causal relationship. However, a critical component in causal learning is active \emph{interaction} with the physical world, based on whether perceived information matches predictions from causal hypotheses. In this work, we combine causal learning (a form of model-building) with a model-based planner to effectively achieve tasks in environments where dynamics are unknown.

In contrast to this work beyond the associative account of causal understanding, recent success in the field of deep \acf{rl} has produced a wide body of research, showcasing agents learning how to play games~\cite{mnih2015human,silver2016mastering,schulman2015trust,schulman2017proximal} and develop complex robotic motor skills~\cite{levine2016end,lillicrap2015continuous} using associative learning schemes. However, the majority of model-free \ac{rl} methods still have great difficulty transferring learned policies to new environments with consistent underlying mechanics but some dissimilar surface features~\cite{zhang2018study,kansky2017schema}. 
This deficiency is due to the limited scope of the agent's overall objective: learning which actions will likely lead to future rewards based on the current state of the environment. 
In traditional \ac{rl} architectures, changes to the location and orientation of critical elements (instance-level) in the agent's environment \emph{appear} as entirely new states, even though their functionality often remains the same (in the abstract-level). Since model-free \ac{rl} agents do not attempt to encode transferable rules governing their environment, new situations appear as entirely new worlds. 
Although an agent can devise expert-level strategies through experiences in an environment, once that environment is perturbed, the agent must repeat an extensive learning process to relearn an effective policy in the altered environment.

In this work, the transfer learning problem is viewed as a combination of instance-level associative learning and abstract-level causal learning. We propose: (i) a bottom-up associative learning scheme that determines which attributes are associated with changes in the environment, and (ii) a top-down causal structure learning scheme that infers which atomic causal structures are useful for a task.
The outcomes of actions are used to update beliefs about the causal hypothesis space, and our agent learns a dynamics model capable of solving our task. 
Specifically, we utilize a virtual ``escape room'' where agents are trapped in an empty room with a locked door. There is a series of conspicuous levers placed around the room with which an agent may interact. Agents placed in such a room may randomly push or pull on the levers to revise their theory about the door's locking mechanism based on observed changes in the environment's state. 
Once an agent discovers a solution, the agent is placed back into the same room but tasked with finding the \emph{next} (different) solution. The agent ``escapes'' from the room after finding \emph{all} of the solutions that can be used to unlock the door.

After completing (escaping) a single room, the agent is placed into a similar room, but with newly positioned levers. Although the levers are in different positions, the rules governing this new room are the same as the last. Thus, the agent's task is to identify the role of each lever, according to the previously learned rules. Because these rules are abstract descriptions of the latent state of the escape room, we refer to the underlying theory as a causal schema~\cite{heider1958psychology}; \ie, a conceptual organization of events identified as cause and effect. Once learned, an agent is able to transfer the learned schema despite different arrangements of levers in the room. Finally, we task agents with transferring knowledge with a different but similar causal schema. The new schema may add additional levers (nodes in a graphical model) or, in a more challenging way, rearrange the structure.

This paper integrates multiple modeling approaches to produce a highly capable agent that can learn causal schemas and transfer knowledge to new scenarios. The contribution of this paper is threefold:
\begin{enumerate}[leftmargin=*,noitemsep,nolistsep]
    \item Learning a bottom-up associative theory that encodes which objects and actions contribute to causal relations;
    \item Learning which top-down atomic causal schemas are solutions, thereby learning generalized abstract task structure;
    \item Integrating the top-down and bottom-up learning scheme with a model-based planner to optimally select interventions from causal hypotheses.
\end{enumerate}

The remainder of this paper is organized as follows:
Section~\ref{sec:openlock} describes the OpenLock task. We present the proposed method of causal theory induction and intervention selection in Section~\ref{sec:causal_theory_induction} and Section~\ref{sec:intervention-selection}, respectively. Section~\ref{sec:results} compares the performance of the proposed model against various \ac{rl} algorithms. Section~\ref{sec:conclusion} concludes the paper with discussions.

\section{OpenLock Task}\label{sec:openlock}

The OpenLock task, originally presented in \citet{edmonds2018human}, requires agents to ``escape'' from a virtual room by unlocking and opening a door. The door is unlocked by manipulating the levers in a particular sequence (see Fig.~\ref{fig:starting-config-and-pushing}a). Each lever can be manipulated using the robotic arm to \emph{push} or \emph{pull} on levers. Only a subset of the levers, specifically grey levers, are involved in unlocking the door (\ie, active levers). White levers are never involved in unlocking the door (\ie, inactive levers); however, this information is not provided to agents. Thus, at the instance-level, agents are expected to learn that grey levers are always part of solutions and white levers are not. Agents are also tasked with finding \emph{all} solutions in the room, instead of a single solution.
 
\begin{figure}[t!]
    \centering
    \includegraphics[width=\linewidth]{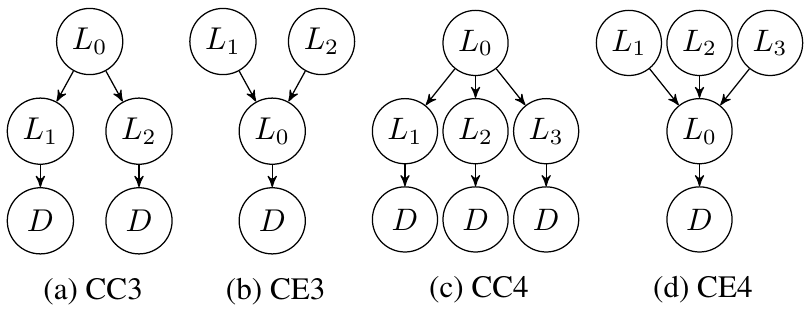}
    \caption{(a) \acf{cc3} causal structure. (b) \acf{ce3} causal structure. (c) \acf{cc4} causal structure. (d) \acf{ce4} causal structure. $L_0$, $L_1$, $L_2$ denote different locks, and $D$ the door.}
    \label{fig:causal-structures}
\end{figure}

\textbf{Schemas:} The door locking mechanism is governed by two causal schemas: \acf{cc} and \acf{ce}. We use the terms \acf{cc3} and \acf{ce3} for schemas with three levers involved in solutions, and \acf{cc4} and \acf{ce4} with four levers; see Fig.~\ref{fig:causal-structures}. Three-lever trials have two solutions; four-lever trials have three solutions. Agents are required to find all solutions within a specific room to ensure that they form either \ac{cc} or \ac{ce} schema structure; a single solution corresponds to a causal chain. 

\textbf{Constraints:} Agents also operate under an action-limit constraint, where only 3 actions (referred to as an \emph{attempt}) can be used to (i) \emph{push} or \emph{pull} on (active or inactive) levers, or (ii) \emph{push} open the door. This action-limit constraint prevents the search depth of interactions with the environment. After 3 actions, regardless of the outcome, the attempt terminates, and the environment resets. Regardless of whether the agent finds all solutions, agents are also constrained to a limited number of attempts in a particular room (referred to as a \emph{trial}; \ie, a sequence of attempts in a room, resulting in finding all the solutions or running out of attempts). An optimal agent will use at most $N+1$ attempts to complete a trial, where $N$ is the number of solutions in the trial. One attempt would be used to identify the role of every lever in the abstract schema, and $N$ attempts would be used for each solution.

\textbf{Training:} Training sessions contain only 3-lever trials. After finishing a trial, the agent is placed in another trial (\ie, room) with the \emph{same} underlying causal schema but with a different arrangement of levers. If agents are forming a useful abstraction of task structure, the knowledge they acquired in previous trials should accelerate their ability to find all solutions in the present and future trials.

\textbf{Transfer:} In the transfer phase, we examine agents' ability to generalize the learned abstract causal schema to \emph{different} but similar environments. We use four transfer conditions consisting of (i) congruent cases where the transfer schema adopts the same structure but with an additional lever (\ac{ce3}-\ac{ce4} and \ac{cc3}-\ac{cc4}), and (ii) incongruent cases where the underlying schema is changed with an additional lever (\ac{cc3}-\ac{ce4} and \ac{ce3}-\ac{cc4}). We compare these transfer results against two baseline conditions (\ac{cc4} and \ac{ce4}), where the agent is trained in a sequence of 4-lever trials.

While seemingly simple, this task is unique and challenging for several reasons. First, requiring the agent to find all solutions rather than a single solution enforces the task as a \ac{cc} or \ac{ce} structure, instead of a single causal chain. Second, transferring the agent between trials with the same underlying causal schema but different lever positions encourages efficient agents to learn an \emph{abstract} representation of the causal schema, rather than learning \emph{instance-level} policies tailored to a specific trial. We would expect agents unable to form this abstraction to perform poorly in any transfer condition. Third, the congruent and incongruent transfer conditions test how well agents are able to adapt their learned knowledge to different but similar causal circumstances. These characteristics of the OpenLock task present challenges for current machine learning algorithms, especially model-free \ac{rl} algorithms.

\section{Causal Theory Induction}\label{sec:causal_theory_induction}

\begin{figure*}[t!]
    \centering
    \includegraphics[width=0.75\linewidth]{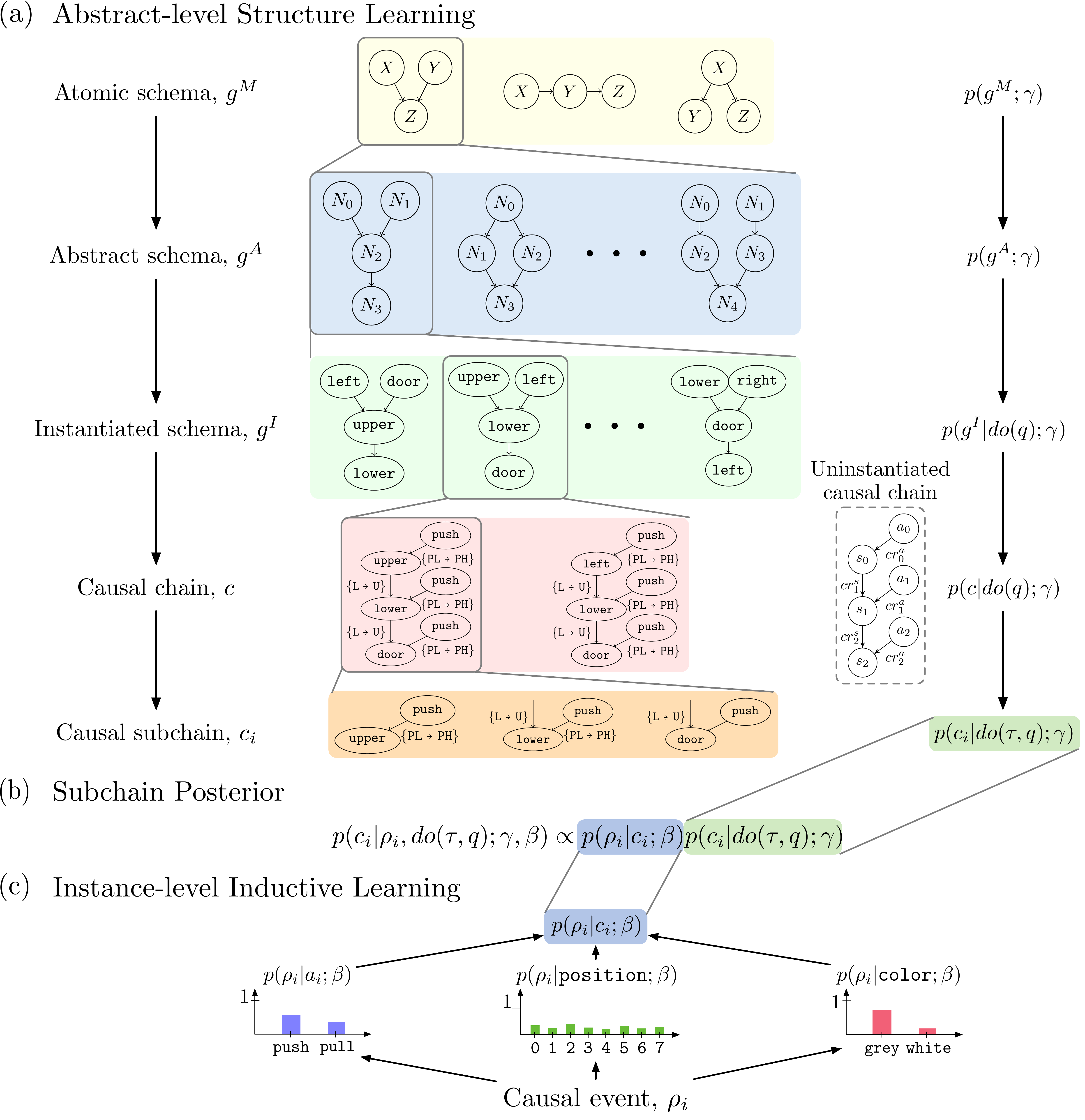}
    \caption{Illustration of top-down and bottom-up processes. (a) Abstract-level structure learning hierarchy. At the top, atomic schemas provide the agent with environment-invariant task structures. At the bottom, causal subchains represent a single time-step in the environment. The agent constructs the hierarchy and makes decisions at the causal subchain resolution. Atomic schemas $g^M$ provide the top-level structural knowledge. Abstract schemas $g^A$ are structures specific to a task, but not a particular environment. Instantiated schemas $g^I$ are structures specific to a task and a particular environment. Causal chains $c$ are structures representing a single attempt; an abstract, uninstantiated causal chain is also shown for notation. Each subchain $c_i$ is a structure corresponding to a single action. \texttt{PL}, \texttt{PH}, \texttt{L}, \texttt{U} denote fluents \emph{pulled}, \emph{pushed}, \emph{locked}, and \emph{unlocked}, respectively. (b) The subchain posterior computed using the abstract-level structure learning and instance-level inductive learning. (c) Instance-level inductive learning. Each likelihood term is learned from causal events, $\rho_i$. Likelihood terms are combined for actions, positions, and colors.}
    \label{fig:hierarchy-and-induction}
\end{figure*}

Causal theory induction provides a Bayesian account of how hierarchical causal theories can be induced from data~\cite{griffiths2005structure,griffiths2009theory,tenenbaum2006theory}. The key insight is: \emph{hierarchy enables abstraction}. At the highest level, a theory provides general background knowledge about a task or environment. Theories consist of principles, principles lead to structure, and structure leads to data. The hierarchy used here is shown in Fig.~\ref{fig:hierarchy-and-induction}a. Our agent utilizes two theories to learn a model of the OpenLock environment: (i) an instance-level associative theory regarding which attributes and actions induce state changes in the environment, denoted as the bottom-up $\beta$ theory, and (ii) an abstract-level causal structure theory about which atomic causal structures are useful for the task, denoted as the top-down $\gamma$ theory.

\textbf{Notation, Definition, and Space:}
A hypothesis space, $\Omega_C$, is defined over possible causal chains, $c\in\Omega_C$. Each chain is defined as a tuple of subchains: $c = (c_0, \ldots, c_k)$, where $k$ is the length of the chain, and each subchain is defined as a tuple $c_i= (a_i, s_i, cr^a_i, cr^s_i)$. Each $a_i$ is an action node that the agent can execute, $s_i$ is a state node, $cr^a_i$ is a causal relation that defines how a state $s_i$ transitions under an action $a_i$, and $cr^s_i$ is a causal relation that defines how state $s_i$ is affected by changes to the previous state, $s_{i-1}$. Each $s_i$ is defined by a set of time-invariant \emph{attributes}, $\phi_i$ and time-varying \emph{fluents}, $f_i$~\cite{thielscher1998introduction,maclaurin1742treatise,newton1736method}; \ie, $s_i=(\phi_i, f_i)$. Action nodes can be directly intervened on, but state nodes cannot. This means an agent can directly influence (\ie, execute) an action, but how the action affects the world must be \emph{actively} learned. The structure of the general causal chain is shown in the uninstantiated causal chain in Fig.~\ref{fig:hierarchy-and-induction}a. As an example using Fig.~\ref{fig:starting-config-and-pushing}a and the first causal chain in the causal chain level of Fig.~\ref{fig:hierarchy-and-induction}a, if the agent executes \emph{push} on the \emph{upper}  lever, the \emph{lower} lever may transition from \emph{pulled} to \emph{pushed}, and the \emph{left} lever may transition from \emph{locked} to \emph{unlocked}.

The space of states is defined as $\Omega_S = \Omega_\phi \times \Omega_F$, where the space of attributes $\Omega_\phi$ consists of position and color, and the space of fluents $\Omega_F$ consists of binary values for lever status (\emph{pushed} or \emph{pulled}) and lever lock status (\emph{locked} or \emph{unlocked}). The space of causal relations is defined as $\Omega_{CR} = \Omega_F \times \Omega_F$, capturing the possibly binary transitions between previous fluent values and the next fluent values. 

State nodes encapsulate both the time-invariant (attributes) and time-varying (fluents) components of an object. Attributes are defined by low-level features (\eg, position, color, and orientation). These low-level attributes provide general background knowledge about how specific objects change under certain actions; \eg, which levers can be pushed/pulled.

\textbf{Method Overview:}
Our agent induces instance-level knowledge regarding which objects (\ie, instances) can produce causal state changes through interaction (see Section~\ref{sec:instance_learning}) and simultaneously learns an abstract structural understanding of the task (\ie, schemas; see Section~\ref{sec:abstract-structure-learning}). The two learning mechanisms are combined to form a causal theory of the environment, and the agent uses this theory to reason about the optimal action to select based on past experiences (\ie, interventions; see Section~\ref{sec:intervention-selection}). After taking an action, the agent observes the effects and updates its model of both instance-level and abstract-level knowledge.

\subsection{Instance-level Inductive Learning} \label{sec:instance_learning}

The agent seeks to learn which instance-level components of the scene are associated with causal events; \ie, we wish to learn a likelihood term to encode the probability that a causal event will occur. We adhere to a basic yet general associative learning theory: \emph{causal relations induce state changes in the environment, and non-causal relations do not}, referred to as the bottom-up $\beta$ theory. We learn two independent components: attributes and actions, and we assume they are independent to learn a general associative theory, rather than specific knowledge regarding an exact causal circumstance.

We define $\Omega_\phi$, the space of attributes, such as position and color, and learn which attributes are associated with levers that induce state changes in the environment. Specifically, an object is defined by its observable features; \ie, the attributes $\phi$. We also define $\Omega_A$, a set of actions and learn a background likelihood over which actions are more likely to induce a state change. We assume attributes and actions are independent and learn each independently.

Our agent learns a likelihood term for each attribute $\phi_{ij}$ and action $a_i$ using Dirichlet distributions because they serve as a conjugate prior to the multinomial distribution. First, a global Dirichlet parameterized by $\alpha^G$ is used across all trials to encode long-term beliefs about various environments. Upon entering a new trial, a local Dirichlet parameterized by $\alpha^L \in [1,10]$ is initialized to $k\alpha^G$, where $k$ is a normalizing factor. Such design of using a scaled local distribution is necessary to allow $\alpha^L$ to adapt faster than $\alpha^G$ within one trial; \ie, agents must adapt more rapidly to the current trial compared to across all trials. Thus, we have a set of Dirichlet distributions to maintain beliefs: a Dirichlet for each attribute (\eg, position, and color) as well as a Dirichlet for actions. Similarly, we maintain a Dirichlet distribution over each action $a_i$ to encode beliefs regarding which actions are more likely to cause a state change, independent from any particular circumstance.

We introduce $\rho$ to represent a causal event or observation occurring in the environment. Our agent wishes to assess the likelihood of a particular causal chain producing a causal event. The agent computes this likelihood by decomposing the chain into subchains
\begin{equation}
    p(\rho|c;\beta) = \prod_{c_i\in c} p(\rho_i|c_i;\beta),
\end{equation}
where $p(\rho_i|c_i;\beta)$ is formulated as
\begin{equation}
    p(\rho_i|c_i;\beta) \propto p(\rho_i|a_i;\beta) \prod_{\substack{\phi_{ij} \in s_i\\s_i\in c_i}} p(\rho_i|\phi_{ij};\beta),
\end{equation}
where $p(\rho_i|\phi_{ij};\beta)$ and $p(\rho_i|a_i;\beta)$ follow multinomial distributions parameterized by a sample from the attribute and action Dirichlet distribution, respectively.\footnote{See supplementary materials for additional details.\label{footn:supp}} Intuitively, this bottom-up associative likelihood encodes a naive Bayesian prediction of how likely a particular subchain is to be involved with any causal event by considering how frequently the attributes and actions have been in causal events in the past, without regard for task structure. For example, we would expect an agent in OpenLock to learn that grey levers move under certain circumstances and white levers never move. This instance-level learning provides the agent with task-invariant, basic knowledge about which subchains are more likely to produce a causal effect.

\subsection{Abstract-level Structure Learning}\label{sec:abstract-structure-learning}

In this section, we outline how the agent learns abstract schemas; these schemas are used to encode generalized knowledge about task structure that is invariant to a specific observational environment.

A space of atomic causal schemas, $\Omega_{g^M}$, of causal chain, \ac{cc}, and \ac{ce}, serve as categories for the Bayesian prior. The belief in each atomic schema is modeled as a multinomial distribution, whose parameters are defined by a Dirichlet distribution. This root Dirichlet distribution's parameters are updated after every trial according to the top-down causal theory $\gamma$, computed as the minimal graph edit distance between an atomic schema and the trial's solution structure. This process yields a prior over atomic schemas, denoted as $p(g^M;\gamma)$, and provides the prior for the top-down inference process. Such abstraction allows agents to transfer beliefs between the abstract notions of \ac{cc} and \ac{ce} without considering task-specific requirements; \eg, 3- or 4-lever configurations.

Next, we compute the belief in abstract instantiations of the atomic schemas. These abstract schemas share structural properties with atomic schemas but have a structure that matches the task definition. For instance, each schema must have three subchains to account for the 3-action limit imposed by the environment and should have $N$ trajectories, where $N$ is the number of solutions in the trial. Each abstract schema is denoted as $g^A$, and the space of abstract schemas, denoted $\Omega_{g^A}$, is enumerated. The belief in an abstract causal schema is computed as
\begin{equation}
    p(g^A;\gamma) = \sum_{g^M\in\Omega_{g^M}} p(g^A|g^M)p(g^M;\gamma).
\end{equation}
The abstract structural space can be used to transfer beliefs between rooms; however, we need to perform inference over settings of positions and colors \emph{in this trial} as the agent executes. Thus, the agent enumerates a space of instantiated schemas $\Omega_{g^I}$, where each $g^I$ is an instantiated schema. The agent then computes the belief in an instantiated schema as
\begin{equation}
    p(g^I|do(q);\gamma) = \sum_{g^A\in\Omega_{g^A}} p(g^I|g^A,do(q))p(g^A;\gamma),
\end{equation}
where $do(q)$ represents the $do$ operator~\cite{pearl2009causality}, and $q$ represents the solutions already executed. Conditioning on $do(q)$ constrains the space to have instantiated solutions that contain the solutions already discovered by the agent in this trial. Causal chains $c$ define the next lower level in the hierarchy, where each chain corresponds to a single attempt. The belief in a causal chain is computed as
\begin{equation}
    p(c|do(q);\gamma) = \sum_{g^I\in\Omega_{g^I}} p(c|g^I, do(q))p(g^I|do(q);\gamma).
\end{equation}
Finally, the agent computes the belief in each possible subchain as
{\fontsize{9.5}{9.5}\selectfont
\begin{equation}
    p(c_i|do(\tau,q);\gamma) = \sum_{c\in\Omega_C}p(c_i|c,do(\tau,q))p(c|do(q);\gamma),
\end{equation}
}
where $do(\tau,q)$ represents the intervention of performing the action sequence executed thus far in this attempt $\tau$, and performing all solutions found thus far $q$. This hierarchical process allows the agent to learn and reason about abstract task structure, taking into consideration the specific instantiation of the trial, as well as the agent's history within this trial.\footnotemark[\value{footnote}]

Additionally, if the agent encounters an action sequence that does not produce a causal event, the agent prunes all chains that contain the action sequence from $\Omega_C$ and prunes all instantiated schemas that contain the corresponding chain from $\Omega_{g^I}$. This pruning strategy means the agent assumes the environment is deterministic and updates its theory about which causal chains are causally plausible through interactions on-the-fly.

\section{Intervention Selection}\label{sec:intervention-selection}

Our agent's goal is to pick the action it believes has the highest chance of (i) being causally plausible in the environment \emph{and} (ii) being part of the solution to the task. We decompose each subchain $c_i$ into its respective parts, $c_i = (a_i, s_i, cr^a_i, cr^s_i)$. The agent combines the top-down and bottom-up processes into a final subchain posterior:
\begin{equation}
    p(c_i|\rho_i,do(\tau,q);\gamma,\beta) \propto p(\rho_i|c_i;\beta)p(c_i|do(\tau,q);\gamma).
\end{equation}
Next, the agent marginalizes over causal relations and states to obtain a final, action-level term to select interventions:
{\fontsize{8.5}{8.5}\selectfont
\begin{multline}
    p(a_i|\rho_i, do(\tau,q);\gamma, \beta) = \\ \sum_{s_i\in\Omega_{S}}\sum_{cr^a_i\in\Omega_{CR}}\sum_{cr^s_i \in\Omega_{CR}} p(a_i,s_i, cr^a_i, cr^s_i|\rho_i, do(\tau,q);\gamma,\beta).
\end{multline}
}%
The agent uses a model-based planner to produce action sequences capable of opening the door (following human participant instructions in~\cite{edmonds2018human}). The goal is defined as reaching a particular state $s^*$, and the agent seeks to execute the action $a_t$ to maximize the posterior subject to the constraints that the action appears in the set of chains that satisfy the goal, $\Omega_{C^*} = \{c \in \Omega_C~|~s^* \in c\}$. We define the set of actions that appear in chains satisfying the goal as $\Omega_{A^*}=\{a \in \Omega_A | \exists c \in \Omega_{C^*}, \exists~s,cr^a, cr^s~| (a,s,cr^a, cr^s) \in c\}$. The agent's final planning goal is
\begin{equation}
    a_t^* = \underset{a_i \in \Omega_{A^*}}{\argmax}~p(a_i|\rho_i, do(\tau,q);\gamma, \beta).
\end{equation}
At each time-step, the agent selects the action that maximizes this planning objective and updates its beliefs about the world as described in Section~\ref{sec:instance_learning} and Section~\ref{sec:abstract-structure-learning}. This iterative process consists of optimal decision-making based on the agent's current understanding of the world, followed by updating the agent's beliefs based on the observed outcome.

\section{Experiments}\label{sec:results}

We compare results between predominate model-free \ac{rl} algorithms with the proposed theory-based causal transfer model. Specifically, we compare the proposed method against \acf{dqn}~\cite{mnih2015human}, \acf{dqnpe}~\cite{schaul2016prioritized}, \acf{a2c}~\cite{mnih2016asynchronous}, \acf{trpo}~\cite{schulman2015trust}, \ac{ppo}~\cite{schulman2017proximal}, and \acf{maml}~\cite{finn2017model} agents. We use the term \emph{positive transfer} and \emph{negative transfer} to indicate that agent performance benefits from or is hindered by the training phase, respectively.

\subsection{Experimental Setup}

The proposed model follows the same procedure as the one used for human studies presented in \citet{edmonds2018human}. Baseline (no transfer) agents are placed in 4-lever scenarios for all trials. Transfer agents are evaluated in two phases: training and transfer. For every training trial, the agent is placed into a 3-lever trial and allowed 30 attempts to find \emph{all} solutions. In the transfer phase, the agent is tasked with a 4-lever trial. Critically, the agent only sees each trial (room) one time, so generalizations must be formed quickly to transfer between trials successfully. See Section~\ref{sec:openlock} for more details. 

When executing various model-free \ac{rl} agents under this experimental setup, no meaningful learning takes place. Instead, we train \ac{rl} agents by looping through all rooms repeatedly (thereby seeing each room multiple times). Agents are also allowed 700 attempts in each trial to find all solutions. During training, agents execute for 200 training iterations, where each iteration consists of looping through all six 3-lever trials. During transfer, agents execute for 200 transfer iterations, where each iteration consists of looping through all five 4-lever trials. Note that the setup for \ac{rl} agents is advantageous; in comparison, both the proposed model and human subjects are only allowed 30 attempts (versus 700) during the training and 1 iteration (versus 200) for transfer.

\ac{rl} agents operate directly on the state of the simulator encoded as a 16-dimensional binary vector: 
(i) the status of each of the 7 levers (\textit{pushed} or \textit{pulled}), 
(ii) the color of each of the 7 levers (\textit{grey} or \textit{white}), 
(iii) the status of the door (\textit{open} or \textit{closed}) and 
(iv) the status of the door lock indicator (\textit{locked} or \textit{unlocked}). The 7-dimensional encoding of the status and color of each lever encodes the position of each lever; \eg, the 0-th index corresponds to the upper-right position. Despite direct access to the simulator's state, \ac{rl} approaches were unable to form a transferable task abstraction.

Additionally, we utilized a plethora of reward functions to explore under what circumstances these \ac{rl} approaches may succeed. Our agents used sparse reward functions, shaped reward functions, and conditional reward functions that encourage agents to find unique solutions.\footnote{\label{note:rl}See supplementary materials for the numerous architectures, parameters, and reward functions used.} A reward function that only rewards for unique solutions performed best, meaning agents were only rewarded the \textit{first} time they found a particular solution. This is similar to the human experimental setup, under which participants were informed when they found a solution for the first time (thereby making progress towards the goal of finding \textit{all} solutions) but were not informed they executed the same solution multiple times (thereby not making progress towards the goal).

\begin{figure*}[t!]
    \centering
    \includegraphics[width=0.8\linewidth]{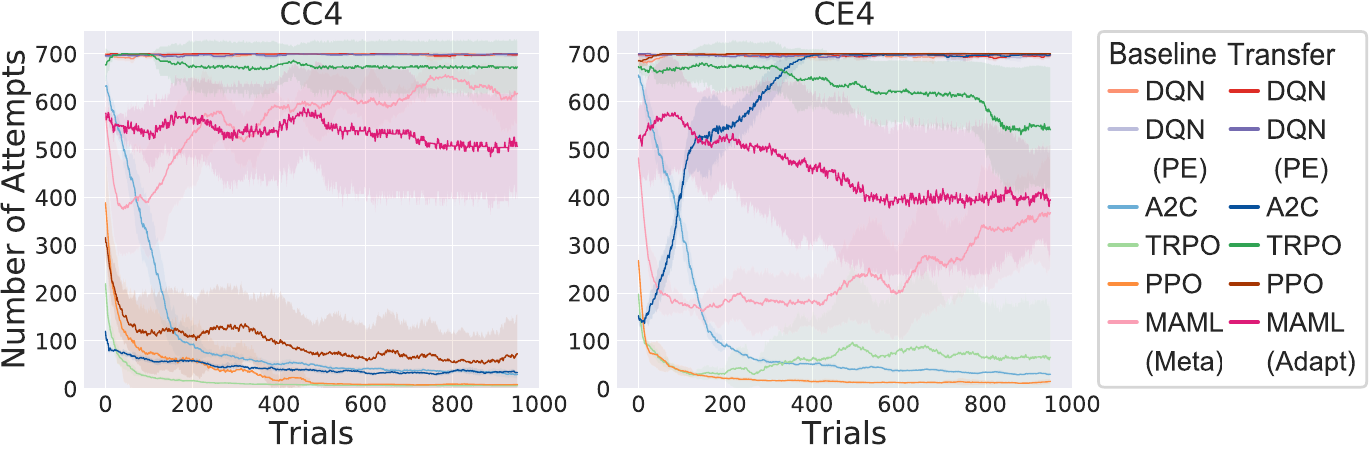}
    \caption{\ac{rl} results for baseline and transfer conditions. Baseline (no transfer) results show the best-performing algorithms (\ac{ppo}, \ac{trpo}) achieving approximately 10 and 25 attempts by the end of the baseline training for \ac{cc4} and \ac{ce4}, respectively. \ac{a2c} is the only algorithm to show positive transfer; \ac{a2c} performed better with training for the \ac{cc4} condition. The last 50 iterations are not shown due to the use of a smoothing function.}
    \label{fig:rl_result_cc4_ce4}
\end{figure*}

\subsection{Reinforcement Learning Results}

The model-free \ac{rl} results, shown in Fig.~\ref{fig:rl_result_cc4_ce4}, demonstrate that \ac{a2c}, \ac{trpo}, and \ac{ppo} are capable of learning how to solve the OpenLock task from scratch. However, \ac{a2c} in the \ac{cc4} condition is the only agent showing positive transfer; every other agent in every condition shows negative transfer.

These results indicate that current model-free \ac{rl} algorithms are capable of learning how to achieve this task; however, the capability to transfer the learned abstract knowledge is markedly different compared to human performance in \citet{edmonds2018human}. Due to the overall negative transfer trends shown by nearly every \ac{rl} agent, we conclude that these \ac{rl} algorithms cannot capture the correct abstractions to transfer knowledge between the 3-lever training phase and the 4-lever transfer phase. Note that the \ac{rl} algorithms found the \ac{ce4} condition more difficult than \ac{cc4}, a result also shown in our proposed model results and human participants.

\subsection{Theory-based Causal Transfer Results}

\begin{figure*}[t!]
    \centering
    \includegraphics[width=\linewidth,]{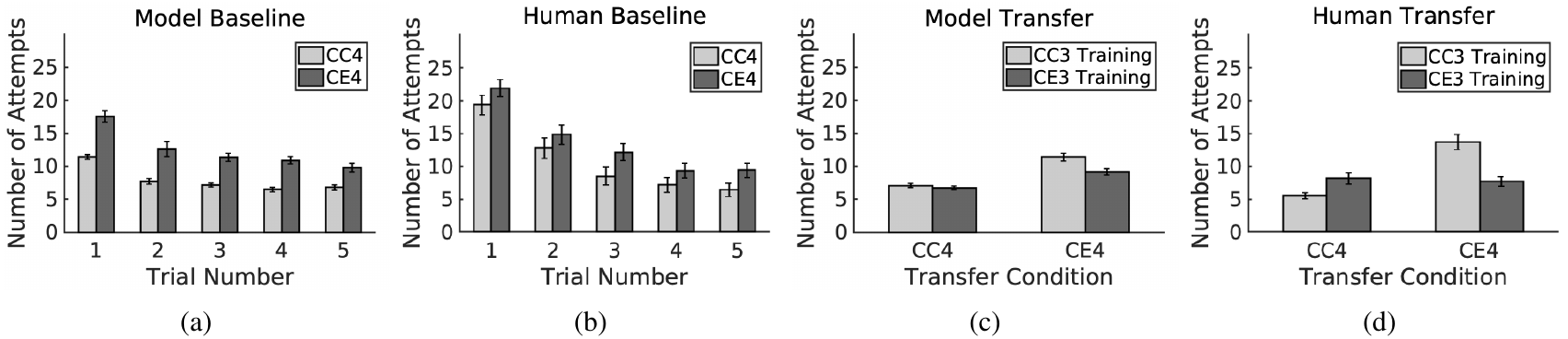}
    \caption{Model performance \emph{vs.} human performance. (a) Proposed model baseline results for \ac{cc4}/\ac{ce4}. We see an asymmetry between the difficulty of \ac{cc} and \ac{ce}. (b) Human baseline performance \protect\cite{edmonds2018human}. (c) Proposed model transfer results for training in \ac{cc3}/\ac{ce3}. The transfer results show that transferring to an incongruent CE4 condition (\ie, different structure, additional lever; \ie, \ac{cc3} to \ac{ce4}) was more difficult than transferring to a congruent condition (\ie, same structure, additional lever; \ie, \ac{ce3} to \ac{ce4}). However, the agent did not show a significant difference in difficulty when transferring to congruent or incongruent condition for the \ac{cc4} transfer condition. (d) Human transfer performance \protect\cite{edmonds2018human}.}
    \label{fig:model_results}
\end{figure*}

The results using the proposed model are shown in Fig.~\ref{fig:model_results}. These results are qualitatively and quantitatively similar to the human participant results presented in \citet{edmonds2018human}, and starkly different from the \ac{rl} results. We execute 40 agents in each condition, matching the number of human subjects described in \citet{edmonds2018human}.

Our agent does not require looping over trials multiple times; it is capable of learning and generalizing from seeing each trial only one time. In the baseline agents, the \ac{ce4} condition was more difficult than \ac{cc4}; this trend was also observed in human participants. During transfer, we see a similar performance as the baseline results; however, for congruent cases (transferring from the same structure with an additional lever) were easier than incongruent cases (transferring to a different structure with an additional lever; \ac{ce4} transfer); this result was statistically significant for \ac{ce4}: $t(79) = 3.0$; $p = 0.004$. For \ac{cc4} transfer, no significance was observed ($t(79) = 0.63$; $p= 0.44$), indicating both \ac{cc3} and \ac{ce3} obtained near-equal performance when transferred to \ac{cc4}.

These learning results are significantly different from the \ac{rl} results; the proposed causal theory-based model is capable of learning the correct abstraction using instance and structural learning schemes, showing similar trends as the human participants. It is worth noting that \ac{rl} agents were trained under highly advantageous settings. \ac{rl} agents: (i) were given more attempts per trial; and (ii) more importantly, were allowed to learn in the same trial multiple times. In contrast, the present model learns the proper mechanisms to: (i) transfer knowledge to structurally equivalent but observationally different scenarios (baseline experiments); (ii) transfer knowledge to cases with structural differences (transfer experiments); and (iii) do so using the \emph{same experimental setup} as humans. The model achieves this by understanding which scene components are capable of inducing state changes in the environment while leveraging overall task structure.\footnote{For additional model results and ablations, see supplementary.}

\section{Conclusion and Discussion}\label{sec:conclusion}

In this work, we show how the theory-based causal transfer coupled with an associative learning scheme can be used to learn transferable structural knowledge under both observationally and structurally varying tasks. We executed a plethora of model-free \ac{rl} algorithms, none of which learned a transferable representation of the OpenLock task, even under favorable baseline and transfer conditions. In contrast, the proposed model results are not only capable of successfully completing the task, but also adhere closely to the human participant results in \citet{edmonds2018human}.

These results suggest that current model-free \ac{rl} methods lack the necessary learning mechanisms to learn generalized representations in hierarchical, structured tasks. Our model results indicate human causal transfer follows similar abstractions as those presented in this work, namely learning abstract causal structures and learning instance-specific knowledge that connects this particular environment to abstract structures. The model presented here can be used in any reinforcement learning environment where: (i) the environment is governed by a causal structure, (ii) causal cues can be uncovered from interacting with objects with observable attributes, and (iii) different circumstances share some common causal properties (structure and/or attributes).

\subsection{Discussion}

\noindent\textbf{Why is causal learning important for \ac{rl}?}
We argue that causal knowledge provides a succinct, well-studied, and well-developed framework for representing cause and effect relationships. This knowledge is invariant to extrinsic rewards and can be used to accomplish many tasks. In this work, we show that leveraging abstract causal knowledge can be used to transfer knowledge across environments with similar structure but different observational properties.

\noindent\textbf{How can \ac{rl} benefit from structured causal knowledge?} 
Model-free \ac{rl} is apt at learning a representation to maximize a reward within simple, non-hierarchical environments using a greedy process. Thus, current approaches do not restrict or impose learning an abstract structural representation of the environment. \ac{rl} algorithms should be augmented with mechanisms to learn explicit structural knowledge and jointly optimized to learn both an abstract structural encoding of the task while maximizing rewards. 

\noindent\textbf{Why is \ac{ce} more difficult than \ac{cc}?}
Human participants, \ac{rl}, and the proposed model all found \ac{ce} more difficult than \ac{cc}. A natural question is: why? We posit that it occurs from a decision-tree perspective. In the \ac{cc} condition, if the agent makes a mistake on the first action, the environment will not change, and the rest of the attempt is bound to fail. However, should the agent choose the correct grey lever, the agent can choose either remaining grey levers; both of which will unlock the door. Conversely, in the \ac{ce} condition, the agent has two grey levers to choose from in the first action; both will unlock the lever needed to unlock the door. However, the second action is more ambiguous. The agent could choose the correct lever, but it could also choose the other grey lever. Such complexity leads to more failure paths from a decision-tree planning perspective. The \ac{cc} condition receives immediate feedback on the first action as to whether or not this plan will fail; the \ac{ce} condition, on the other hand, has more failure pathways. We plan to investigate this property further, as this asymmetry was unexpected and unexplored in the literature.

\noindent\textbf{What other theories may be useful for learning causal relationships?}
In this work, we adhere to an associative learning theory. We adopt the theory that \emph{causal relationships induce state changes}. However, other theories may also be appealing. For instance, the associative theory used does not directly account for long-term relationships (delayed effects). More complex theories could potentially account for delayed effects; \eg, when an agent could not find a causal attribute for a particular event, the agent could examine attributes jointly to best explain the causal effect observed. Prior work has examined structural analogies~\cite{hinrichs2011transfer,zhang2019raven,zhang2019learning} and object mappings~\cite{fitzgerald2018human} to facilitate transfer; these may also be useful to acquire transferable causal knowledge.

\noindent\textbf{How can hypothesis space enumeration be avoided?}
Hypothesis space enumeration can quickly become intractable as problems increase in size. While this worked used a fixed, fully enumerated hypothesis space, future work will include examining how sampling-based approaches can be used to iteratively generate causal hypotheses.~\citet{bramley2017formalizing} showed a Gibbs-sampling based approach; however, this sampling should be guided with top-down reasoning to guide the causal learning process by leveraging already known causal knowledge with proposed hypotheses.

\noindent\textbf{How well would model-based \ac{rl} perform in this task?}
Model-based \ac{rl} may exhibit faster learning within a particular environment but still lacks mechanisms to form abstractions that enable human-like transfer. This is an open research question, and we plan on investigating how abstraction can be integrated with model-based \ac{rl} methods.

\noindent\textbf{How is this method different from hierarchical \ac{rl}?}
Typically, hierarchical \ac{rl} is defined on a hierarchy of goals, where subgoals represent \emph{options} that can be executed by a high-level planner~\cite{chentanez2005intrinsically}. Each causally-plausible hypothesis can be seen as an option to execute. This work seeks to highlight the importance of leveraging causal knowledge to form a world-model and using said model to guide a reinforcement learner. In fact, our work can be recast as a form of hierarchical model-based \ac{rl}.

Future work should primarily focus on how to integrate the proposed causal learning algorithm directly with reinforcement learning. An agent capable of integrating causal learning with reinforcement learning could generalize world dynamics (causal knowledge) and goals (rewards) to novel but similar environments. One challenge, unaddressed in this paper, is to how to generalize rewards to varied environments. Traditional reinforcement learning methods, such as Q-learning, do not provide a mechanism to extrapolate internal values to similar but different states. In this work, we showed how extrapolating causal knowledge can aid in uncovering the causal relationships in similar environments. Adopting a similar scheme for some form of reinforcement learning would enable reinforcement learners to succeed in the OpenLock task without iterating over the trials multiple times, and could enable one-shot reinforcement learning. Future work will also examine how a learner can iteratively grow a causal hypothesis while incorporating a background theory of causal relationships.

\section*{Acknowledgement}

The authors thank Chi Zhang at the UCLA Computer Science Department, Feng Gao, Prof. Tao Gao, and Prof. Ying Nian Wu at the UCLA Statistics Department for helpful discussions. This work reported herein is supported by MURI ONR N00014-16-1-2007, DARPA XAI N66001-17-2-4029, ONR N00014-19-1-2153, and an NVIDIA GPU donation grant.

\bibliographystyle{aaai}
\fontsize{9.8}{9.8}\selectfont
\bibliography{bibliography}

\end{document}